%% file: main.tex
\documentclass[10pt,twocolumn,letterpaper]{article}

\usepackage[pagenumbers]{cvpr} %

\input{preamble}

\definecolor{cvprblue}{rgb}{0.21,0.49,0.74}
\usepackage[pagebackref,breaklinks,colorlinks,citecolor=cvprblue]{hyperref}

\def\paperID{*****} %
\def\confName{CVPR}
\def\confYear{2024}

\title{
Beyond Pixels: Exploring Human-Readable SVG Generation for Simple Images with Vision Language Models

}

\author{Tong Zhang\\
University of California Irvine  \\
{\tt\small tongz27@uci.edu}
\and
Haoyang Liu\\
University of Illinois at Urbana-Champaign \\
{\tt\small hl57@illinois.edu}
\and 
Peiyan Zhang\\
Hong Kong University of Science and Technology\\
{\tt\small zhangao@cse.ust.hk}
\and 
Yuxuan Cheng\\
Huazhong Agricultural University\\
{\tt\small hxwxss@webmail.hzau.edu.cn}
\and
Haohan Wang\\
University of Illinois at Urbana-Champaign\\
{\tt\small haohanw@illinois.edu}
}

\begin{document}
\maketitle
\input{sec/0_abstract}    
\input{sec/1_intro}

\input{sec/2_methods}
\input{sec/3_exp}

{
    \small
    \bibliographystyle{ieeenat_fullname}
    \bibliography{main}
}

\end{document}

%% file: preamble.tex
\usepackage{algorithm}
\usepackage{amssymb}
\usepackage{algpseudocode}
\usepackage[dvipsnames]{xcolor}

\newcommand\tsup[1]{\raisebox{-0.5ex}{\textsuperscript{#1}}}

%% file: sec/0_abstract.tex
\begin{abstract}

In the field of computer graphics, the use of vector graphics, particularly Scalable Vector Graphics (SVG), represents a notable development from traditional pixel-based imagery. SVGs, with their XML-based format, are distinct in their ability to directly and explicitly represent visual elements such as shape, color, and path. This direct representation facilitates a more accurate and logical depiction of graphical elements, enhancing reasoning and interpretability.
Recognizing the potential of SVGs, the machine learning community has introduced multiple methods for image vectorization. However, transforming images into SVG format while retaining the relational properties and context of the original scene remains a key challenge. 
Most vectorization methods often yield SVGs that are overly complex and not easily interpretable.
In response to this challenge, we introduce our method, Simple-SVG-Generation (S\textsuperscript{2}VG\textsuperscript{2}). 
Our method focuses on producing SVGs that are both accurate and simple, aligning with human readability and understanding. 

With simple images, we evaluate our method with reasoning tasks 
together with advanced language models, the results show a clear improvement over previous SVG generation methods. 
We also conducted surveys for human evaluation on the readability of our generated SVGs, the results also favor our methods. 

\end{abstract}

%% file: sec/1_intro.tex
\section{INTRODUCTION}
Rival image representation techniques coexist within computer graphics: bitmap images, which consist of pixel matrices, and vector images, depict sequences of artistically drawn shapes. Image Rasterization is particularly well understood and implemented, yet modern-day trends show a rising interest towards image vectorization for its inherent advantages, such as scale adaptability and resolution independence, which are vital for usage in contemporary interfaces~\cite{dziuba2023image}. However, vectorization poses numerous challenges, particularly in generating vector images or Scalable Vector Graphics (SVGs)~\cite{w3c01svg} that are both human-readable and retain the semantic relevance of the original raster image.

\begin{figure}[t]

\centering
\includegraphics[width=1.0\linewidth]{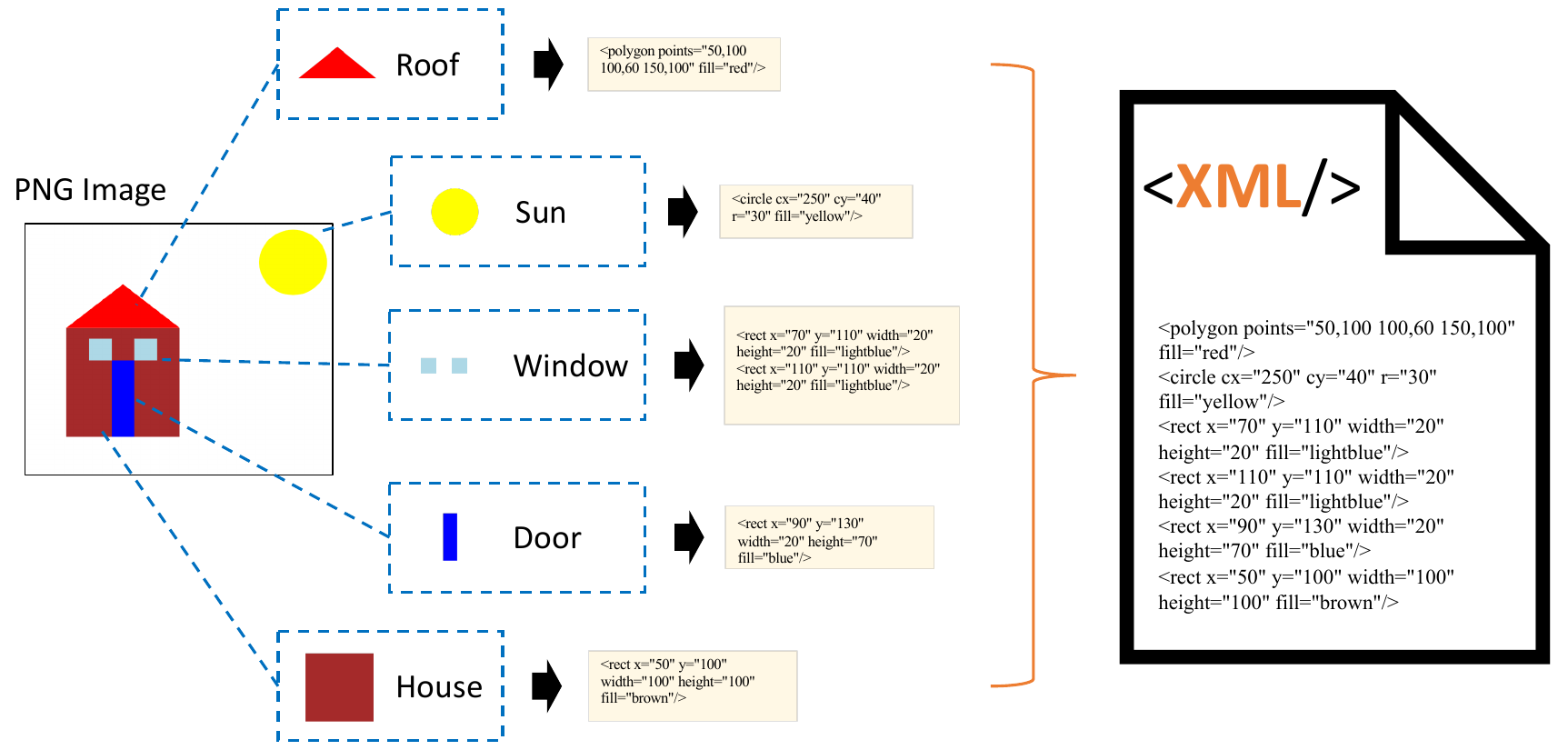}
\caption{Demonstrates an SVG representation of a simplified house, translating basic geometrical shapes into their corresponding SVG code elements and then into a visual structure, highlighting the vectorization process. }
\label{fig:opening}
\end{figure}

Contrasting raster images, which are an assembly of ordered pixels, SVGs describe images through a set of parametric shape primitives which enables numerous benefits including smaller file sizes and resolution-independence~\cite{10.1109/MMUL.2003.1218261}. Despite being inherently more compact and flexible, ensuring that the SVG representations retain the semantics of the original image and at the same time are efficiently generable remains a difficult task. Figure \ref{fig:opening} provides a visual representation of the fundamental components of an SVG. It illustrates how simple geometric shapes are coded and rendered into a composite image. 

Recent years have seen considerable progress in image vectorization, primarily through two technical advancements: implementing advanced generative models~\cite{carlier2020deepsvg,DBLP:journals/corr/abs-1904-02632} and deploying sophisticated differentiable rendering methods~\cite{Li:2020:DVG,xu2022live}. However, these developments overlooked key characteristics of SVGs, such as their XML foundation and their complexity beyond simple paths. Specifically, these methods typically yield SVGs that are unreadable, as they overly rely on certain strategies like the DiffVG, which generates SVGs with a fixed number of paths where many are unnecessarily obscured by others~\cite{Li:2020:DVG}. Moreover, these images frequently contain curves that overshoot the viewBox attribute's boundaries, resulting in a vectorized image that fails to capture the intended semantic information. These methods also spends quite a long time to generate SVGS for image preprocessing and optimization process. Therefore, a simple but practical method for capturing the essence of image vectorization is sought after in the field.

In this paper, we propose a novel approach, S\textsuperscript{2}VG\textsuperscript{2}, which leverages the capabilities of vision language models to overcome these challenges. S\textsuperscript{2}VG\textsuperscript{2} shows excellent performance in generating clean, understandable SVG representations from complex raster images, while maintaining their meaning from a comprehensible, human perspective. Efficacy of S\textsuperscript{2}VG\textsuperscript{2} is evaluated rigorously by benchmarking it against three key metrics: pixel and feature level similarity to the original bitmap; simplicity and readability of the resulting image in terms of the number of shapes and the parameters defining them; and the speed at which the vectorized image is generated. 

S\textsuperscript{2}VG\textsuperscript{2}'s ability to encapsulate the semantic structure of intricate raster images into a neatly arranged collection of elementary shape primitives represents a crucial stride in the field of image vectorization aided by vision language models. This work paves the way for new methodologies in creating SVG icons and explores the potential of combining vision and language models for complex graphical tasks, thereby opening new possibilities in computer graphics and image processing fields.

Moreover, it poses an innovative perspective on vectorization process by incorporating vision language models. S\textsuperscript{2}VG\textsuperscript{2} unravels a new realm of possibilities for pretrained vision large language models in visual representation and understanding. Our main contributions in this work can be summarized as follows:
\begin{itemize}
   \item  We propose S\textsuperscript{2}VG\textsuperscript{2}, a novel method combined with a vision language model for intricate SVG generation, which can effectively retain the semantic coherence from raster images to SVGs and is capable of generating human-readable SVGs.
    \item S\textsuperscript{2}VG\textsuperscript{2} is extensively benchmarked and evaluated against performance metrics involving the vision quality of image vectorization, readability for language models and simplicity of the final SVG  over previous methods.
    \item  We introduce a specialized dataset named SVG-SHAPE, designed for evaluating SVG generation methods. Each image in this dataset is a 384×384 RGB image depicting a 3×3 grid of objects, providing a standardized and challenging testbed for assessing vectorization algorithms.
\end{itemize}

\section{RELATED WORK}
\subsection{Scalable Vector Graphics}
Vector graphics provide an alternative representation framework for visual content, delineating images as collections of parameterized shape primitives, such as polygons, circles, and rectangles~\cite{Peng2004TheRO}
. This contrasts with conventional raster-based images composed of pixel grids. Notably, vector graphics leverage these primitives' geometric attributes, represented by coordinate sets defining contours and associated color values. This representation is widely supported by web browsers, requiring no specialized software or plugins for rendering. The succinctness and infinite scalability of vector graphics result in adaptability via easy adjustments in stroke or color parameters.

The Scalable Vector Graphics format (SVG) takes advantage of this intrinsic characteristic of vector graphics by encoding images as XML-based text files, specifying geometric entities and their pertinent attributes. This XML structure empowers facile manipulation and editing, rendering SVG particularly versatile for web applications and graphic design tasks. In the context of this paper, we harness the potential of vision language models (VLMs)~\cite{zhang2023visionlanguage} to bridge the gap between pixel-based images and the SVG format. This strategic integration of VLMs ensures the fidelity of generated SVGs in terms of readability, simplicity, and preservation of semantic coherence.

\subsection{Image Vectorization}
Rasterization and vectorization are dual problems in
image processing. Various techniques, including algorithmic and machine learning-based methods~\cite{tian}, have been explored. Algorithmic approaches encompass mesh-based~\cite{73b5eb95f5354a718e4c4e16f7d795eb,liao,Zhou2014RepresentingIU} and curve-based methods~\cite{7962259,6745296}, each with their own advantages and limitations. Machine learning-based methods, however, offer promising avenues for automated vector image generation~\cite{dziuba2023image} .

Existing machine learning-compatible vectorization methods, such as Mang2Vec~\cite{su2023marvel} and DVoTD~\cite{Egiazarian_2020}, exhibit limitations. Mang2Vec struggles with color images and often generates SVGs with an excessive number of shapes, impacting both efficiency and semantic clarity.  Other methods, like DiffVG~\cite{Li:2020:DVG} and LIVE~\cite{xu2022live}, leverage iterative processes and optimization techniques, offering varying levels of accuracy and efficiency.
\subsection{Image Captioning }
Image captioning~\cite{vinyals2015tell} is the task of generating a coherent natural language description for an image, representing a vibrant and actively researched intersection of computer vision and natural language processing in recent years. The central objective of image captioning lies in the development of models capable of learning the intricate mapping between visual and textual domains, with the aim of producing meaningful and precise image descriptions.

In contemporary approaches to image captioning, notable efforts have been channeled towards elevating the quality of generated captions. These endeavors encompass the integration of attention mechanisms, reinforcement learning techniques, and the utilization of pretrained vision language models. Attention mechanisms~\cite{mnih2014recurrent} empower models to selectively focus on specific regions within an image during the caption generation process. Reinforcement learning~\cite{mnih2013playing}, on the other hand, equips models with the capacity to refine their captioning skills by learning from feedback received during training.

Recent advancements have introduced large language models (LLMs) like GPT-4~\cite{openai2023gpt4}, which have showcased remarkable multi-modal capabilities. A prominent research focus has been the alignment of image encoders with these pretrained vision language models~\cite{liu2023llava,zhu2023minigpt,kim2021vilt}, representing a compelling category within the broader domain of pretrained vision language models (VLMs). These pretrained VLMs have the ability to encapsulate extensive vision-language correspondence knowledge, facilitating zero-shot predictions through the matching of embeddings derived from both images and texts.

%% file: sec/2_methods.tex
\section{Methodology}

In this section, we present the intricacies of the S\textsuperscript{2}VG\textsuperscript{2} method, which leverages the combined strengths of vision and language models to enhance the generation of Scalable Vector Graphics (SVG). We explore the architecture that forms the backbone of our approach in Section~\ref{sec:arc}, and later detail our strategic training approach designed to fine-tune and refine this architecture for optimal SVG generation in Section~\ref{sec:loss}.

\subsection{Architecture}
\label{sec:arc}
\begin{figure*}[h]
\centering
\includegraphics[width=1.0\linewidth]{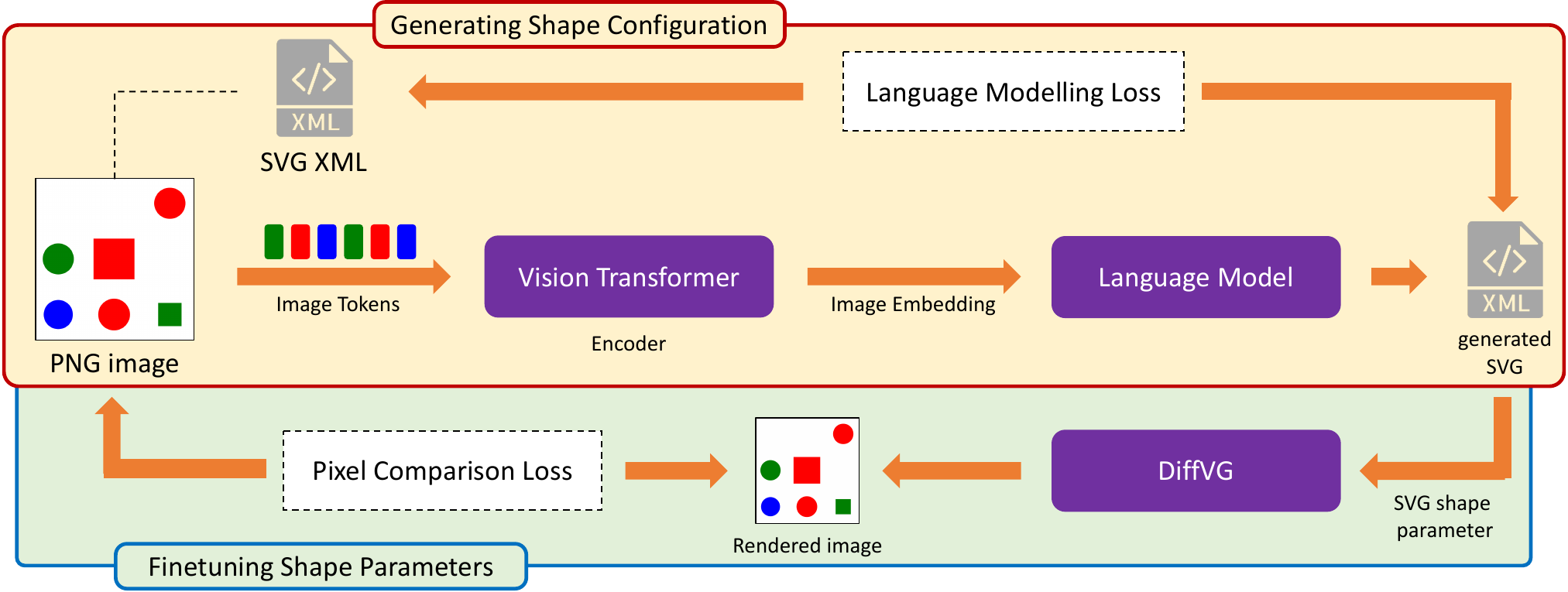}
\caption{Overview of the Proposed S\textsuperscript{2}VG\textsuperscript{2} Approach.}
\end{figure*}

The architecture of our S\textsuperscript{2}VG\textsuperscript{2} system is designed to simplify and enhance the flexibility of SVG generation, employing the capabilities of pre-trained vision-language models. While our method is generally applicable to vision-language models, we use the BLIP~\cite{li2022blip} model in our work for its efficiency and widespread usage, to facilitate experiments and benchmarking. The model integrates a vision transformer~\cite{dosovitskiy2021image} as a visual encoder and a pre-trained BERT~\cite{devlin2019bert} model with causal attention for language generation, forming a framework that leverages the strengths of both visual perception and language semantics. The visual transformer processes the image with its attention mechanism to efficiently capture the global context, and a BERT model adapted with causal attention is used as the text decoder, which conducts cross-attention to integrate information from the vision encoder, and generate tokens of the SVG code auto-regressively.

\subsection{Loss Function}
\label{sec:loss}
Our primary objective is defined as training a machine learning model that predicts Scalable Vector Graphics (SVG) code from a given raster image. The generated SVG code should accurately restore the original image and concisely represent its semantic information, which is essential for reasoning tasks.

\paragraph{Training Stage}
We leverage the strong ability of the pretrained vision-language model in capturing the joint distribution of image and text, and fine-tune it on our SVG dataset to transfer its knowledge to the image vectorization task.
In this stage, we align the model's prediction of the SVG code given the image with the ground truth SVG code. 

Let $D$ denote the dataset of paired images and labels $\{(\mathbf{x}_i, y_i)\}_{i=1}^N$, where $\mathbf{x}_i$ is the raster image and $y_i$ is the SVG code. $y_i$ is a sequence of $l_i$ tokens, $y_i = (y_i^{(1)}, y_i^{(2)}, \ldots, y_i^{(l_i)})$. The task can be formulated as estimating the probability of the next token conditional on the raster image and the previous tokens. 
\begin{align}
P(y_i^{(t)} | y_i^{(1)}, \ldots, y_i^{(t-1)}, \mathbf{x}_i)
\end{align}

In our model, the input image $\mathbf{x}_i$ is processed by a vision transformer, which divides the image into $M$ patches and encodes them as a sequence of embeddings $\mathbf{V}_i = \{\mathbf{v}_i^{(1)}, \mathbf{v}_i^{(2)}, ..., \mathbf{v}_i^{(M)}\}$, including a [CLS] token for the global image feature.

The model's image-grounded text decoder the takes embeddings of the previous tokens $(\mathbf{y}_i^{(1)}, \ldots, \mathbf{y}_i^{(t-1)})$ as input, and conduct cross attention over the visual tokens, to estimate the probability of the next token
\begin{equation}
  \begin{aligned}
\hat{P}(y_i^{(t)} | y_i^{(1)}, \ldots, y_i^{(t-1)}, \mathbf{x}_i) &= \\
f_{\theta}(y_i^{(t)} | \mathbf{v}_i^{(1)}, ..., & \mathbf{v}_i^{(M)}, \mathbf{y}_i^{(1)}, \ldots, \mathbf{y}_i^{(t-1)})
 \end{aligned}
\end{equation}

where $f_{\theta}(\cdot | \cdot)$ denotes the conditional generative model, and $\theta$ is its parameters. Our model is fine-tuned using the cross-entropy loss, averaged over all timesteps and samples in the dataset, as shown in the below equation:
\begin{align}
\mathcal{L(\theta)} = \sum_{i=1}^N \sum_{t=1}^{l_i} - \log \hat{P}(y_i^{(t)} | y_i^{(1)}, \ldots, y_i^{(t-1)}, \mathbf{x}_i)
\end{align}

\paragraph{Inference Stage}

For text generation, the model's text decoder operates autoregressively, predicting the probability of all candidate tokens, and choose the one with the highest probability as the next token, to get the predicted SVG code $\hat{y}_i$.

Using the general knowledge in the pretrained model and the above fine-tuning process, our model can give good predictions on the shape configuration of the images, leading to decent performance in our experiments. However, the shape parameters predicted are by the model are probably not ideal, because large models are known to be less strong in quantitative reasoning . To improve the generation quality, we further conduct a step during the inference stage to refining 

We use regular expressions to parse the SVG code $\hat{y}_i$ into a set of shapes and corresponding parameters, denoted as $S_i$, $S_i = \{(s_i^{(j)}, \phi_i^{(j)})\}_{j=1}^{r_i}$, where $r_i$ is the number of objects in the predicted SVG code, $s_i^{(j)}$ denotes the shape type of the $j$-th object, chosen from pre-defined categories, and $\phi_i^{(j)}$ is the parameter that specifies its exact shape, size, and layout in the image.

We take $S_i$ as the input to DiffVG for rendering the image. DiffVG perform the rendering of image as a differentiable process, such that the pixels of the rendered image are differentiable w.r.t the shape parameters. This results in rendered image $\tilde{\mathbf{x}}_i$. This builds a computation graph integreted in auto-differentition frameworks. 
\begin{equation}
    \tilde{\mathbf{x}}_i = \operatorname{DiffVG}(S_i)
\end{equation}
After rendering the initial SVG code into an image, we optimize the shape parameters with gradient descent, based on the similarity between this rendered image and the actual test image. The loss function is shown below.
\begin{align}
    \mathcal{L}({\phi}) = \alpha \cdot ||\mathbf{x}_i - \tilde{\mathbf{x}}_i||_1 + \beta \cdot ||\mathbf{x}_i - \tilde{\mathbf{x}}_i||_2
\end{align}

Here, \( \alpha \) and \( \beta \) are weights for the $\ell_1$ and $\ell_2$ norms, respectively. This loss is used to refine the parameters of the SVG to better match the original image.

The described method demonstrates the efficacy of combining a vision-language model with DiffVG rendering for SVG generation. This approach not only captures the semantic structure of the input images but also refines the SVG output to ensure high fidelity and semantic coherence.

%% file: sec/3_exp.tex
\section{Experiments}
\subsection{Datasets}
In this section, we provide a detailed description of the experimental datasets utilized to evaluate the performance of our S\textsuperscript{2}VG\textsuperscript{2} method. We introduce two distinct datasets: SVG-SHAPE and SVG-Transform.
\begin{figure}[htbp]
\centering
\begin{subfigure}[b]{0.2\textwidth}
    \includegraphics[width=\textwidth]{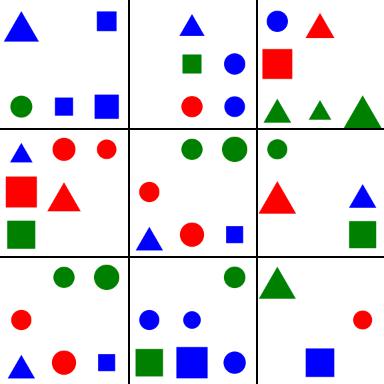}
    \caption{SVG-SHAPE}
    \label{fig:shape}
\end{subfigure}\hfill
\begin{subfigure}[b]{0.2\textwidth}
    \includegraphics[width=\textwidth]{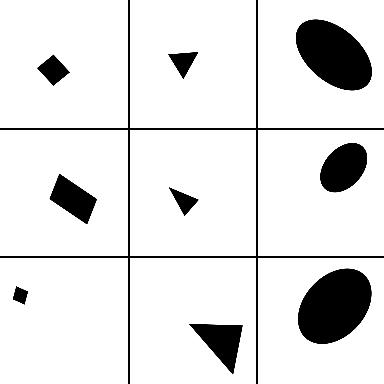}
    \caption{SVG-Transform }
    \label{fig:transform}
\end{subfigure}\hfill

\caption{Images  samples from the SVG-SHAPE and SVG-Transform datasets. }
\label{fig:example}
\end{figure}
\subsubsection{SVG-SHAPE Dataset}
To thoroughly assess the capabilities of our S\textsuperscript{2}VG\textsuperscript{2} method, we have constructed a novel dataset, named SVG-SHAPE. This dataset, depicted in Figure~\ref{fig:shape}, is designed to challenge and evaluate the performance of SVG generation algorithms in visual question answering contexts. Details of the dataset construction are provided in Algorithm~\ref{alg:svgshape}.

\begin{algorithm}
\caption{Construction of the SVG-SHAPE Dataset}
\label{alg:svgshape}
\begin{algorithmic}[1]
\State Initialize an empty dataset $D$ for SVG-SHAPE

\For{$i$ in Dataset Size}
    \State Initialize an empty SVG document $svg$
    \For{$j$ in Number of Grids}
        \State Generate a random shape (circle, rectangle, or triangle)
        \State Generate a random color (red, green, or blue)
        \State Generate a random size for the object
        \State Create an SVG element for the object using the generated shape, color, and size
        \State Append the SVG element to the $svg$ document
    \EndFor
    \State Add the generated $svg$ to the dataset $D$
\EndFor
\State Rasterize all the SVGs  into  RGB images

\end{algorithmic}
\end{algorithm}

The SVG-SHAPE dataset is specifically crafted for visual question answering tasks. It features images where diverse shapes, varying in color and size, are strategically placed on a 3x3 grid. Each image is associated with two types of tasks: one that requires inferring the presence of specific shapes, for example, "Is there a green circle?", and another that involves determining relative positions, such as "Is there a red triangle positioned above a blue shape?"

\subsubsection{SVG-Transform Dataset}
To enhance the capabilities in SVG generation and manipulation, we present the SVG-Transform dataset. This dataset, showcased in Figure~\ref{fig:transform}, is designed to challenge and assess SVG generation techniques in a more intricate scenario: performing geometric transformations on shapes. Details of the dataset construction are provided in Algorithm~\ref{alg:svgtransform}.

\begin{algorithm}
\caption{Construction of the SVG-Transform Dataset}
\label{alg:svgtransform}
\begin{algorithmic}[1]
\State Initialize an empty dataset $D$ for SVG-Transform

\For{$i$ in Dataset Size}
    \State Initialize an empty SVG document $svg$
    \State Generate a random shape (circle, rectangle, or triangle)
    \State Generate a random size for the object
    \State Generate a random combination transform (translate, scale, skewX, skewY and rotate)
    \State Create an SVG element for the object using the generated shape, color, and size
    \State Append the SVG element to the $svg$ document
    \State Add the generated $svg$ to the dataset $D$
\EndFor
\State Rasterize all the SVGs  into  RGB images

\end{algorithmic}
\end{algorithm}

\subsection{Training Settings}

In our experimental setup, we conduct a comparative analysis between our innovative method and established baseline techniques in the realm of SVG generation. Training involves two distinct datasets: the SVG-SHAPE dataset, which includes 50,000 image-SVG pairs, and the more comprehensive SVG-Transform dataset, encompassing 500,000 pairs. These datasets provide a solid foundation for evaluating the efficacy and adaptability of S\textsuperscript{2}VG\textsuperscript{2}.

The architectural backbone of our model integrates the Vision Transformer (ViT) with BERT parameters adapted from the BLIP model. This model is pre-trained on a diverse array of datasets, including the Visual Genome (VG)~\cite{krishna2016visual}, the SBU Captions Dataset~\cite{Ordonez:2011:im2text}, and the extensive Conceptual Captions 12M dataset~\cite{changpinyo2021conceptual}. This pre-training equips our model with a nuanced understanding of complex visual-textual relationships.

For the optimization process, we utilize the Adam optimizer, setting the learning rate to 1e-5 and the weight decay coefficient at 0.05. Training is conducted on a powerful computational setup comprising four NVIDIA A40 GPUs, and we maintain a batch size of 16. 
\subsection{Evaluation}
\subsubsection{Image Quality}
We evaluate the vectorization quality of S\textsuperscript{2}VG\textsuperscript{2} through both quantitative and qualitative analysis, focusing on the differences between the input targets and the SVG rendered images.

\begin{figure}[htbp]
\centering

\begin{subfigure}[b]{0.1\textwidth}
    \includegraphics[width=\textwidth]{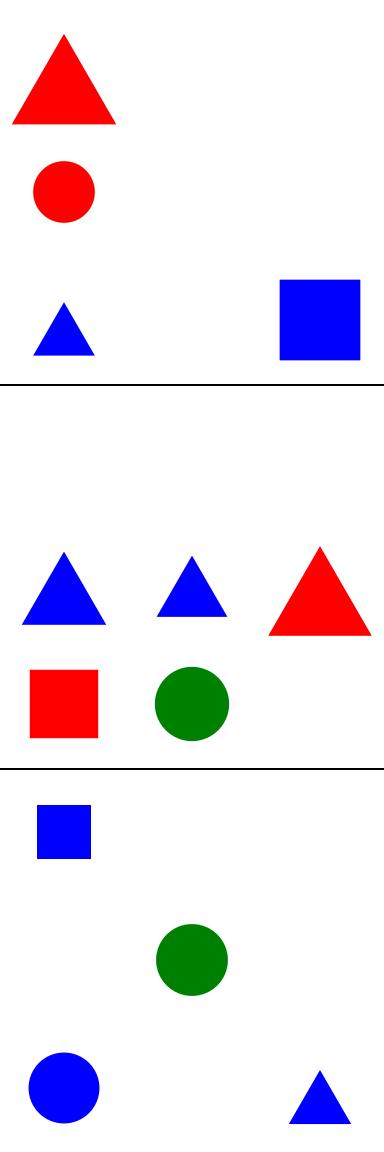}
    \caption{Input}
    \label{fig:2a}
\end{subfigure}\hfill
\begin{subfigure}[b]{0.1\textwidth}
    \includegraphics[width=\textwidth]{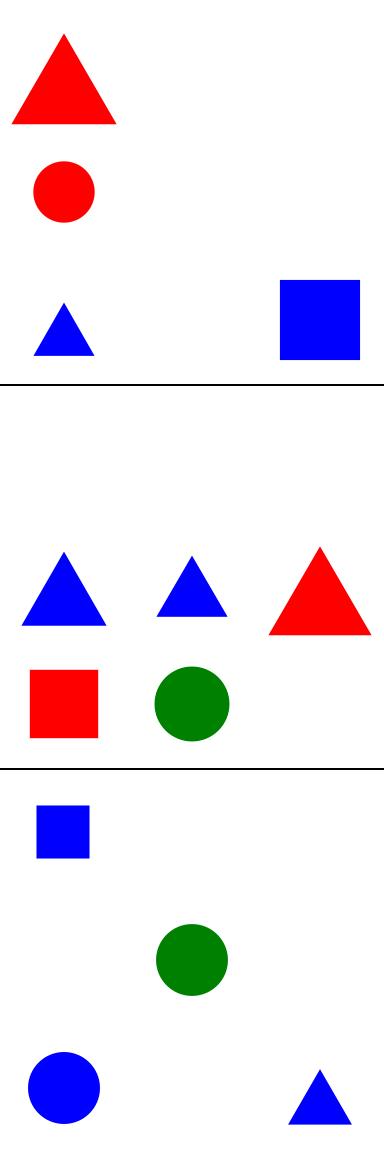}
    \caption{S\textsuperscript{2}VG\textsuperscript{2}}
    \label{fig:2b}
\end{subfigure}\hfill
\begin{subfigure}[b]{0.1\textwidth}
    \includegraphics[width=\textwidth]{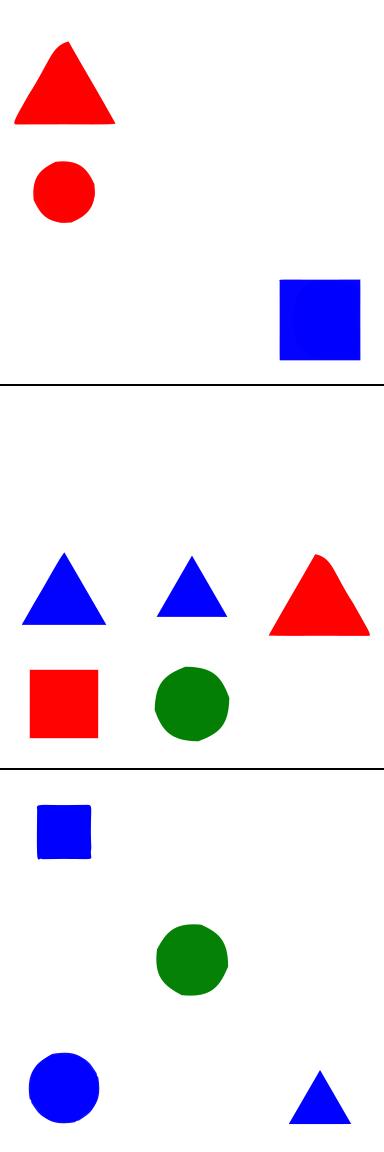}
    \caption{LIVE}
    \label{fig:2c}
\end{subfigure}\hfill
\begin{subfigure}[b]{0.1\textwidth}
    \includegraphics[width=\textwidth]{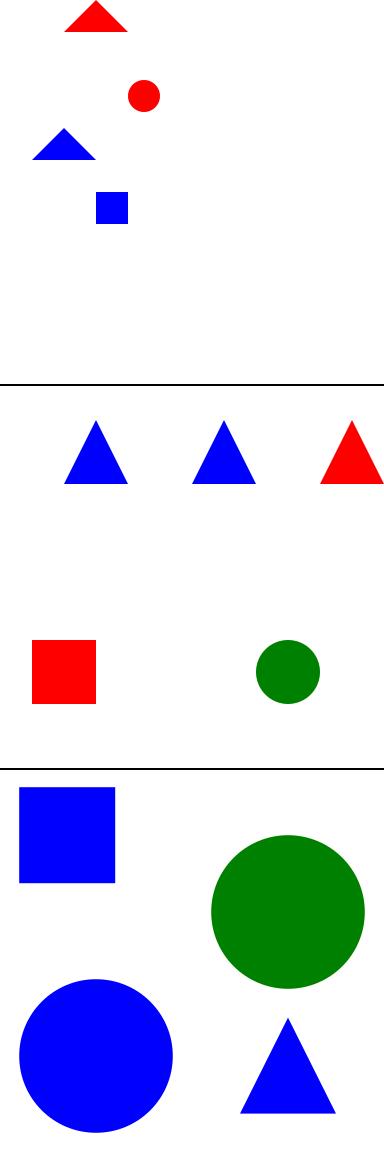}
    \caption{GPT-4V}
    \label{fig:2d}
\end{subfigure}

\caption{Images from different image vectorization methods.
}
\label{fig:2all}
\end{figure}

\begin{figure}[h]

\centering

\includegraphics[width=1.0\linewidth]{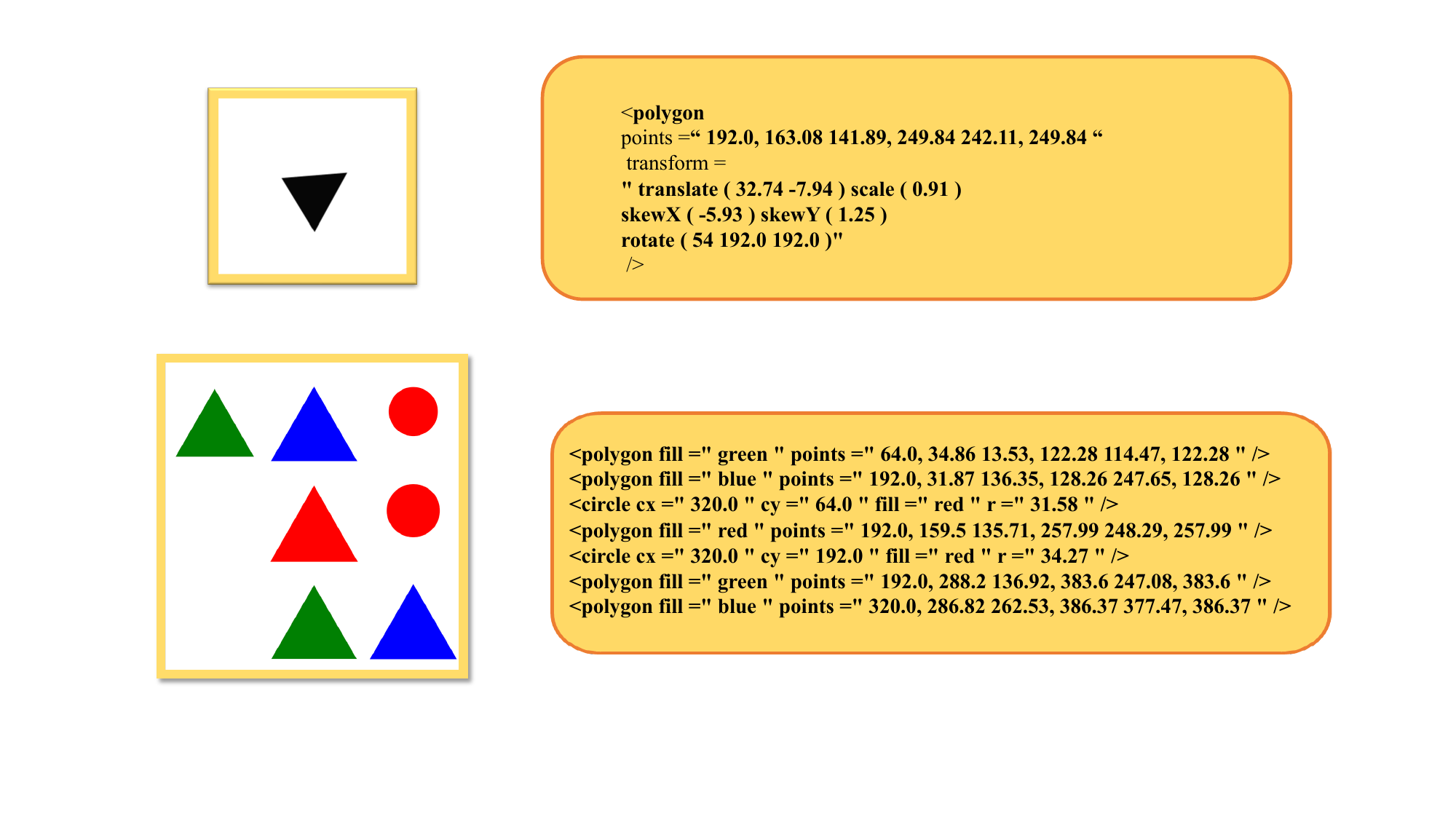}
\caption{The top image showcases a shape from the SVG-Transform dataset, with its SVG code generated by S\textsuperscript{2}VG\textsuperscript{2} , highlighting the human-readable format and precise transformation parameters. The bottom image, from the SVG-SHAPE dataset, presents an array of geometric shapes in vibrant colors. The accompanying SVG code, also produced by S\textsuperscript{2}VG\textsuperscript{2}, clearly outlines the shapes and their attributes, emphasizing the method's effectiveness in generating interpretable and semantically coherent SVGs.}
\label{fig:infer}
\end{figure}

The baseline for comparison is established using the original input images (Figure~\ref{fig:2a}). SVGs generated by S\textsuperscript{2}VG\textsuperscript{2} (Figure~\ref{fig:2b}) exhibit high fidelity to these originals, accurately preserving shapes and colors. In comparison, images vectorized by the LIVE method (Figure~\ref{fig:2c}), with a predetermined path number of 10 and 50 iterations per layer (path), display noticeable distortions in shape geometry and color hues. The vectorization results from GPT-4V (Figure~\ref{fig:2d}) show the most significant deviations in shape accuracy and color representation.  It can only reliably identify some relative positional relationships between objects within the image. Additionally, we vectorize images using DiffVG, setting the path number to 50 and iterating 500 times, a configuration chosen to balance the complexity of SVG tokens against each method's ability to render the images accurately.The readable SVG code produced by S\textsuperscript{2}VG\textsuperscript{2}  for both the SVG-Transform and SVG-SHAPE datasets can be seen in Figure \ref{fig:infer} , illustrating the method's effectiveness in generating interpretable and semantically coherent SVGs.

This visual comparison highlights the superior performance of S\textsuperscript{2}VG\textsuperscript{2} in preserving the integrity of the original designs during vectorization. It is important to note that the specific settings for LIVE and DiffVG, particularly in terms of path number and iteration count, play a crucial role in this evaluation. 

To more precisely quantify the quality of the vectorized images, we employ several image quality metrics for comparison: Learned Perceptual Image Patch Similarity (LPIPS)\cite{zhang2018perceptual}, Structural Similarity Index (SSIM)\cite{article}, as well as L1 and L2 norms. These metrics offer a comprehensive evaluation, shedding light on the perceptual and structural similarities, along with the error magnitudes, between the generated images and their original counterparts.

\begin{table}[h!]
\centering
\begin{tabular}{lcccc}
\hline
\textbf{Method} & \textbf{LPIPS\tsup{$\downarrow$}} & \textbf{SSIM\tsup{$\uparrow$}} & \textbf{L1\tsup{$\downarrow$}} & \textbf{L2\tsup{$\downarrow$}} \\
\hline

S\textsuperscript{2}VG\textsuperscript{2}  & 0.00395 & 0.99501 & 0.92041 & 0.63227 \\
DiffVG & 0.06075 & 0.97949 & 8.66592 & 3.39910 \\
LIVE & 0.01835 & 0.99418 & 15.30974 & 0.79269 \\
\hline
\end{tabular}
\caption{Comparison of Image Quality Metrics for Different Methods }
\label{tab:image_quality_metrics}
\end{table}

The results in Table~\ref{tab:image_quality_metrics} reveal that S\textsuperscript{2}VG\textsuperscript{2} surpasses other methods in all these metrics. Notably, it records the lowest score in LPIPS, suggesting a higher perceptual resemblance to the original images. In terms of structural integrity, it achieves the highest SSIM score. Furthermore, its lower values in both L1 and L2 norms indicate a minimal pixel-level error throughout the vectorization process.

\subsection{Readability Evaluation}
\subsubsection{Visual Question Answering}
\begin{figure}[h]
\centering
\includegraphics[width=1.0\linewidth]{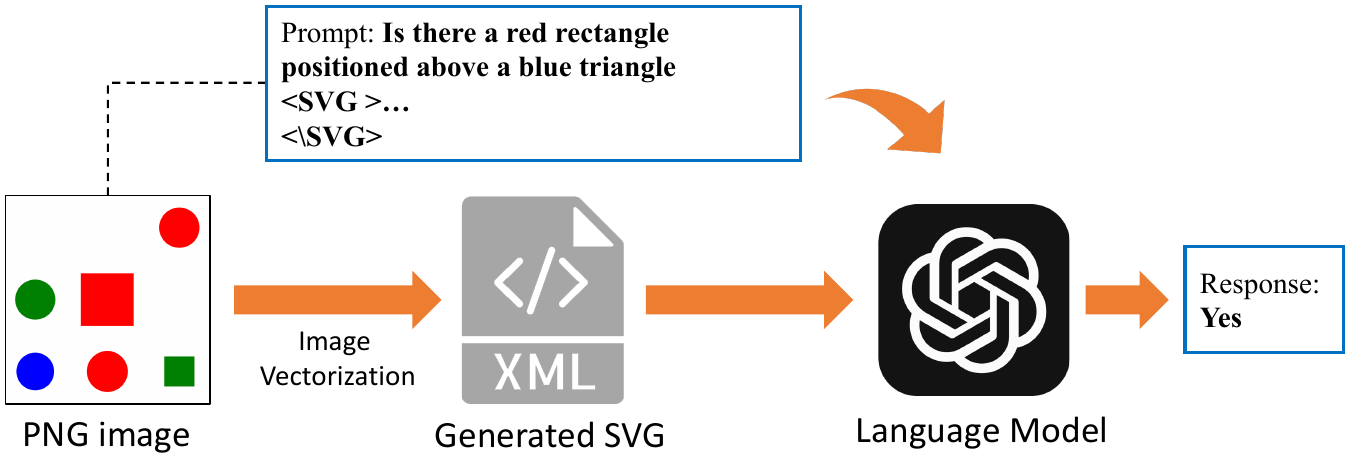}
\caption{Evaluation of the SVG Readability.}
\end{figure}
Visual Question Answering (VQA) is a task that combines image understanding and language generation to answer questions about images. In the context of our work, VQA is employed to assess the readability of the generated SVG.  As shown in Figure~\ref{fig:2all}, the process starts with the conversion of raster images into SVG format using S\textsuperscript{2}VG\textsuperscript{2}. These SVGs, coupled with relevant questions, are then fed into large language models (LLMs) for VQA. The LLM's performance in answering these questions provides insight into the quality of the SVGs in terms of their readability and fidelity to the original images.

In our evaluation framework, we leverage GPT-4-32k, a variant of the GPT-4 model designed to handle inputs with extended token lengths up to 32,000 tokens. This adaptation is crucial as it accommodates the lengthy and complex SVGs typically generated by existing vectorization methods, which often exceed the standard token limits of conventional language models.

\begin{table}[h!]
\centering
\begin{tabular}{lccc}
\hline
\textbf{Method} & \textbf{acc} & \textbf{acc1} & \textbf{acc2} \\
\hline
GT      & 87.725 & 99.25 & 76.2  \\
S\textsuperscript{2}VG\textsuperscript{2}    & 87.75  & 99.05 & 76.45 \\
LIVE    & 50.075 & 50.1  & 50.05 \\
DiffVG  & 50.0   & 50.0  & 50.0  \\
\hline
\end{tabular}
\caption{Accuracy metrics for two VQA tasks. Acc1 represents the accuracy for identifying the existence of specific shapes within the SVG, while Acc2 relates to the accuracy of determining the relative position of shapes.}
\label{tab:acc_vqa}
\end{table}

As shown in Table~\ref{tab:acc_vqa}, our method demonstrates superior readability performance. Its accuracy (acc) closely matches the Ground Truth (GT), the optimal benchmark in this context. Particularly in identifying the existence of specific shapes (acc1), our method reaches near-perfect accuracy. 

In tasks involving the assessment of relative shape positioning (acc2), our method significantly outperforms the LIVE and DiffVG methods. This highlights S\textsuperscript{2}VG\textsuperscript{2}'s effectiveness in preserving spatial relationships and contextual integrity of the original images during conversion.

In summary, the SVGs produced by S\textsuperscript{2}VG\textsuperscript{2} exhibit high readability in VQA tasks, attesting to the method’s ability to generate SVGs that are visually precise and semantically detailed.

\subsubsection{User Study}

We conduct a user study to assess the effectiveness of our SVG generation method, S\textsuperscript{2}VG\textsuperscript{2}, in comparison to other methods like LIVE and DiffVG. The study involves a total of 19 questions, divided into two parts.

The first part includes 9 questions focused on evaluating the participants' ability to infer the existence and relative positions of shapes within the SVGs. The objective is to determine how accurately users interpret and understand the spatial arrangement and presence of shapes in the SVGs. The second part comprises 10 questions where participants are asked to select the SVG output from each method that they find easiest to understand.

\begin{figure}[htbp]
\centering
\begin{subfigure}[b]{0.2\textwidth}
    \includegraphics[width=\textwidth]{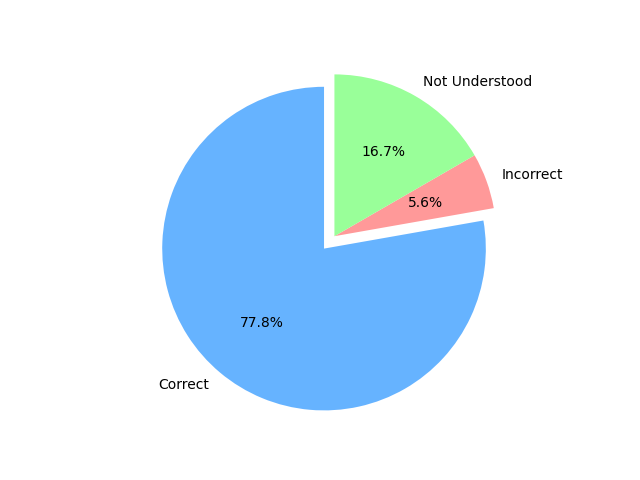}
    \caption{Participant responses for S\textsuperscript{2}VG\textsuperscript{2}, showing a majority correctly identified the SVG representations.}
    \label{fig:3a}
\end{subfigure}
\begin{subfigure}[b]{0.2\textwidth}
    \includegraphics[width=\textwidth]{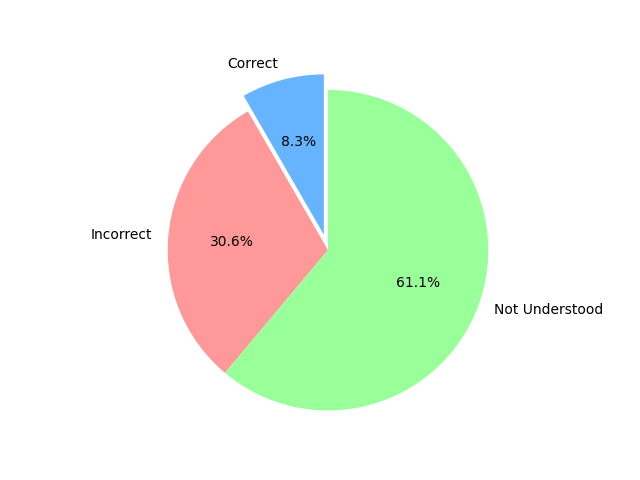}
     \caption{Participant responses for LIVE, indicating a majority found the SVGs difficult to understand.}
    \label{fig:3b}
\end{subfigure}

\begin{subfigure}[b]{0.2\textwidth}
    \includegraphics[width=\textwidth]{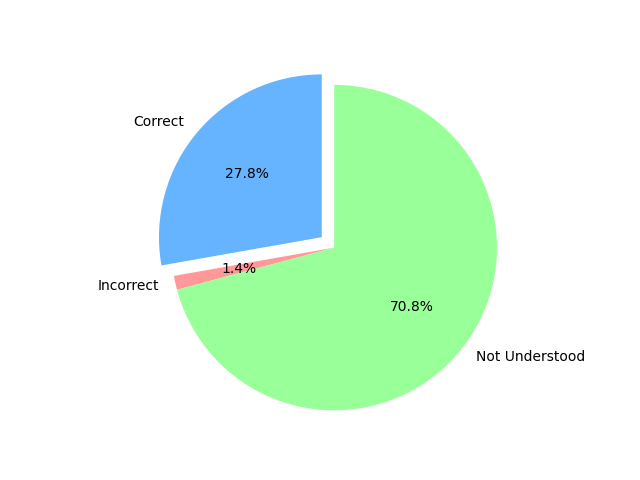}
    \caption{Participant responses for DiffVG,indicating a higher difficulty level to understand compared to LIVE. }
    \label{fig:3c}
\end{subfigure}
\begin{subfigure}[b]{0.2\textwidth}
    \includegraphics[width=\textwidth]{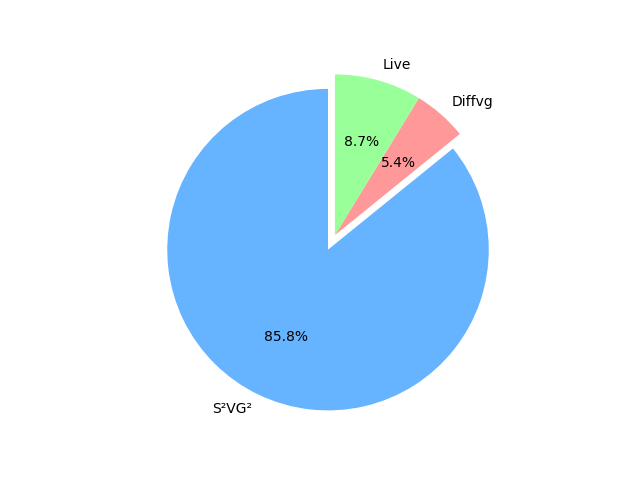}
    \caption{Overall participant preference for the most understandable SVG output, with a significant leaning towards S\textsuperscript{2}VG\textsuperscript{2}.}
    \label{fig:3d}
\end{subfigure}

\caption{Images from different image vectorization methods.}
\label{fig:3all}
\end{figure}
The user study, as shown in Figure~\ref{fig:3all}, provides clear evidence of user preferences in SVG readability. A majority of 77.8\% could easily interpret SVGs generated by S\textsuperscript{2}VG\textsuperscript{2}, as per Figure~\ref{fig:3a}, indicating a strong alignment with human cognitive processing.

In stark contrast, Figure~\ref{fig:3b} shows that for the LIVE method, the majority (61.1\%) could not understand the SVGs, and a significant portion (30.6\%) answered incorrectly, indicating a considerable gap in the clarity and readability of the SVGs produced by this method. Figure~\ref{fig:3c} for DiffVG reflects a slightly better comprehension than LIVE, with 70.8\% of responses indicating a lack of understanding. This could be attributed to the higher complexity and possibly less intuitive vector representations generated by DiffVG. Crucially, Figure~\ref{fig:3d} demonstrates a compelling preference among participants for S\textsuperscript{2}VG\textsuperscript{2} when choosing the easiest to understand SVG output. A significant 85.8\% favored S\textsuperscript{2}VG\textsuperscript{2}, with the remaining percentages dispersed between the other two methods. This overwhelming preference underscores the superior readability and user-friendly nature of the SVGs generated by S\textsuperscript{2}VG\textsuperscript{2}.

The results  from the user study conclusively show that S\textsuperscript{2}VG\textsuperscript{2} significantly enhances the readability of SVGs. They are more comprehensible compared to the alternatives, likely due to the reduced complexity and smaller file sizes of the SVGs generated by our method.

\subsubsection{Complexities Reflected in SVG File Characteristics}
To further our understanding of the practical implications of SVG generation methods, we delve into the complexities as reflected in the file characteristics. We consider the file size and tokens number as a proxy for the complexity of the SVGs, which also has a direct impact on the interpretability and computational efficiency of these vector images.

\begin{table}[h!]
\centering
\begin{tabular}{lcc}
\toprule
\textbf{Method} & \textbf{Mean Size (bytes)} & \textbf{Std Dev (bytes)} \\
\midrule
S\textsuperscript{2}VG\textsuperscript{2} & 390.507 & 112.957 \\
LIVE & 6776.223 & 148.000 \\
DiffVG & 28354.357 & 196.729 \\
\bottomrule
\end{tabular}
\caption{Comparison of Mean File Sizes and Standard Deviations}
\label{tab:file_sizes}
\end{table}

\begin{table}[h!]
\centering
\begin{tabular}{lcc}
\toprule
\textbf{Method} & \textbf{Mean Length (tokens)} & \textbf{Std Dev (tokens)} \\
\midrule
S\textsuperscript{2}VG\textsuperscript{2} & 186.72 & 50.394 \\
LIVE & 3624.951 & 86.126 \\
DiffVG & 15736.102 & 131.635 \\
\bottomrule
\end{tabular}
\caption{Comparison of Mean Token Lengths and Standard Deviations}
\label{tab:token_lengths}
\end{table}
As depicted in Tables \ref{tab:file_sizes} and \ref{tab:token_lengths}, S\textsuperscript{2}VG\textsuperscript{2} demonstrates a clear advantage in generating more concise and less complex SVG files. This is evidenced by the significantly smaller mean file sizes and shorter token lengths. Such streamlined files suggest a higher degree of readability and simplified vector representations, which could lead to more efficient computation and easier user interpretation.

Aligned with the outcomes of our user study, the SVGs from S\textsuperscript{2}VG\textsuperscript{2} were preferred for their clarity and simplicity, underscoring the critical nature of an optimized vectorization process. In essence, the ability to generate SVGs that are both computationally efficient and user-friendly is paramount, particularly when scaling up for broader application in the field of computer graphics and beyond.

\section{Discussion}

This section delves into the implications of our findings from the S\textsuperscript{2}VG\textsuperscript{2} approach, highlighting its strengths, acknowledging potential limitations, and identifying opportunities for future research.

\subsection{Implications of Findings}

The S\textsuperscript{2}VG\textsuperscript{2} model has shown considerable prowess in generating simplified yet semantically rich SVGs. Its success is partly attributed to the innovative use of a pre-trained Vision Transformer and a BERT model with causal attention mechanisms. This approach facilitates the production of SVGs that closely resemble the original images, making it potentially valuable in fields like graphic design automation and image editing.

Quantitatively, the model distinguishes itself in the realm of image vectorization, achieving lower LPIPS scores and higher SSIM values relative to existing methods. These metrics are reflective of the model's adeptness in maintaining visual details and structural integrity, aspects that are vital for the practical application of SVGs in various domains.

\subsection{Enhanced Image Understanding via SVG}

The S\textsuperscript{2}VG\textsuperscript{2} framework showcases the immense potential of SVGs as a medium for advanced image comprehension. It transforms raster images into SVGs, encapsulating detailed semantic information in a structured, vectorized format. This structured representation enables discrete manipulation of image components, thereby facilitating various tasks such as image editing, graphic design, and automated visual content analysis. The efficacy of this approach is rooted in the SVG's ability to depict complex images through scalable, editable, and descriptively rich vector paths. This capability distinctly sets SVGs apart from traditional bitmap images, which rely on fixed pixel grids.

\subsection{Limitations}

While S\textsuperscript{2}VG\textsuperscript{2} marks significant progress in SVG generation, it is important to acknowledge its inherent limitations, which could serve as focal points for future enhancements:

\begin{itemize}
    \item \textbf{Token Limitation}: The current implementation of S\textsuperscript{2}VG\textsuperscript{2} is constrained by a fixed token limit, which restricts its ability to process highly complex images. When dealing with images rich in details and elements, this limitation can lead to an oversimplification of the SVG output and result in a loss of nuanced details or an inability to fully capture the intricacy of the original raster image.

    \item \textbf{Challenges with Direct Backpropagation}: The method encounters difficulties in directly backpropagating visual loss parameters through the network. While we employ differentiable rendering techniques, such as DiffVG, for an indirect approach to optimizing SVG parameters, the lack of a direct backpropagation pathway limits the model's precision. Enhancing this aspect could significantly improve the fidelity of the SVG outputs to their original images, making the vectorization process more accurate and efficient.

    \item \textbf{Handling Complex Images}: One of the notable limitations of S\textsuperscript{2}VG\textsuperscript{2} is its performance with images that possess high levels of complexity. This includes images with intricate patterns, textures, or a large number of distinct visual elements. The current model may struggle to accurately and effectively vectorize such images, which can be critical for applications requiring detailed graphical representations.

\end{itemize}

These limitations highlight areas for future research and development, underscoring the need for ongoing innovation and refinement in the field of SVG generation.

\section{Conclusion}

In this work, we introduce S\textsuperscript{2}VG\textsuperscript{2}, a novel approach for generating Scalable Vector Graphics (SVG) through the integration of vision-language models. S\textsuperscript{2}VG\textsuperscript{2} skillfully navigates the complexities of image vectorization, creating SVGs that are not only accurate but also preserve the semantic essence of the original images. This method marks a notable advance in making SVGs more understandable and manageable, enhancing their interpretability for human users.

Through rigorous experiments and comparative analyses, S\textsuperscript{2}VG\textsuperscript{2} is proven to significantly contribute to the field of simple SVG generation. It exemplifies the successful integration of vision and language models in tackling complex graphical tasks. Our findings open new pathways for exploration in SVG generation, hinting at exciting possibilities for automating graphic design and enhancing tools for visual reasoning.